\documentclass[journal,10pt, onecolumn]{IEEEtran}

\IEEEoverridecommandlockouts                             

\usepackage{amsmath}
\usepackage{algorithm}  
\usepackage{algpseudocode}
\usepackage{multirow}
\usepackage{amsmath}
\usepackage{amssymb}
\usepackage{epsfig} 
\usepackage{amsmath} 
\usepackage{graphicx}
\usepackage{caption}
\usepackage{subcaption}
\usepackage[flushleft]{threeparttable}
\usepackage{tabularx} 

\title{\LARGE \bf
	Enhanced Visual SLAM based Collision-free Driving for Lightweight Autonomous Cars
}

\author{Zhihao Lin$^{1}$\textdagger,  Zhen Tian$^{1}$\textdagger, Qi Zhang$^{2}$, Hanyang Zhuang$^{3}$, and Jianglin Lan$^{1*}$
	\thanks{*This work was supported in part by the China Scholarship Council Ph.D. Scholarship for 2023-2027 (No.202206170011), in part by the Leverhulme Trust Early Career Fellowship (ECF-2021-517), and in part by the UK Royal Society International Exchanges Cost Share Programme (IEC$\backslash$NSFC$\backslash$223228). }
	\thanks{$^{1}$Zhihao Lin, Zhen Tian, and Jianglin Lan are with James Watt School of Engineering, University of Glasgow, Glasgow G12 8QQ, United Kingdom}%
	\thanks{$^{2}$Qi Zhang is with School of Computing Science, University of Glasgow, Glasgow G12 8QQ, United Kingdom}%
	\thanks{$^{3}$Hanyang Zhuang is with the University of Michigan-Shanghai Jiao Tong University Joint Institute, Shanghai Jiao Tong University, Shanghai, 200240, China}
	\thanks{$^{*}$Corresponding author. Jianglin Lan(e-mail: jianglin.lan@glasgow.ac.uk)}%
	\thanks{$\dagger$ Equal contribution}
}

\begin{document}

	\maketitle
	\pagestyle{empty}  
	\thispagestyle{empty} 

\begin{abstract}
The paper presents a vision-based obstacle avoidance strategy for lightweight self-driving cars that can be run on a CPU-only device using a single RGB-D camera. The method consists of two steps: visual perception and path planning. The visual perception part uses ORBSLAM3 enhanced with optical flow to estimate the car's poses and extract rich texture information from the scene. In the path planning phase,  we employ a method combining a control Lyapunov function and control barrier function in the form of quadratic program (CLF-CBF-QP) together with an obstacle shape reconstruction process (SRP) to plan safe and stable trajectories. To validate the performance and robustness of the proposed method, simulation experiments were conducted with a car in various complex indoor environments using the Gazebo simulation environment. Our method can effectively avoid obstacles in the scenes. The proposed algorithm outperforms benchmark algorithms in achieving more stable and shorter trajectories across multiple simulated scenes.
\end{abstract}



\section{Introduction}
In recent years, there has been an increasing demand for autonomous vehicles in complex indoor environments. Compared to drones, autonomous cars can perform a variety of ground transportation tasks while maintaining good stability. To avoid causing damage to the items being transported, it is crucial for unmanned vehicles to remain stable under all circumstances. Therefore, these vehicles should be equipped with obstacle avoidance algorithms to perform effectively in hazardous situations such as the sudden appearance of moving obstacles, tight corners, and narrow corridors. However, the existing algorithms \cite{9973345, 10016694, 9990588} face various practical issues. Unmanned vehicles are often limited by heavy sensors (Radar, LiDAR, etc.), which can lead to short battery life, high costs, and large sizes, limiting the use cases and performance of these algorithms. Thus, choosing the most suitable sensors for unmanned vehicles is a critical task.
\begin{figure}[thpb]   
\centering
\includegraphics[width=0.7\columnwidth]{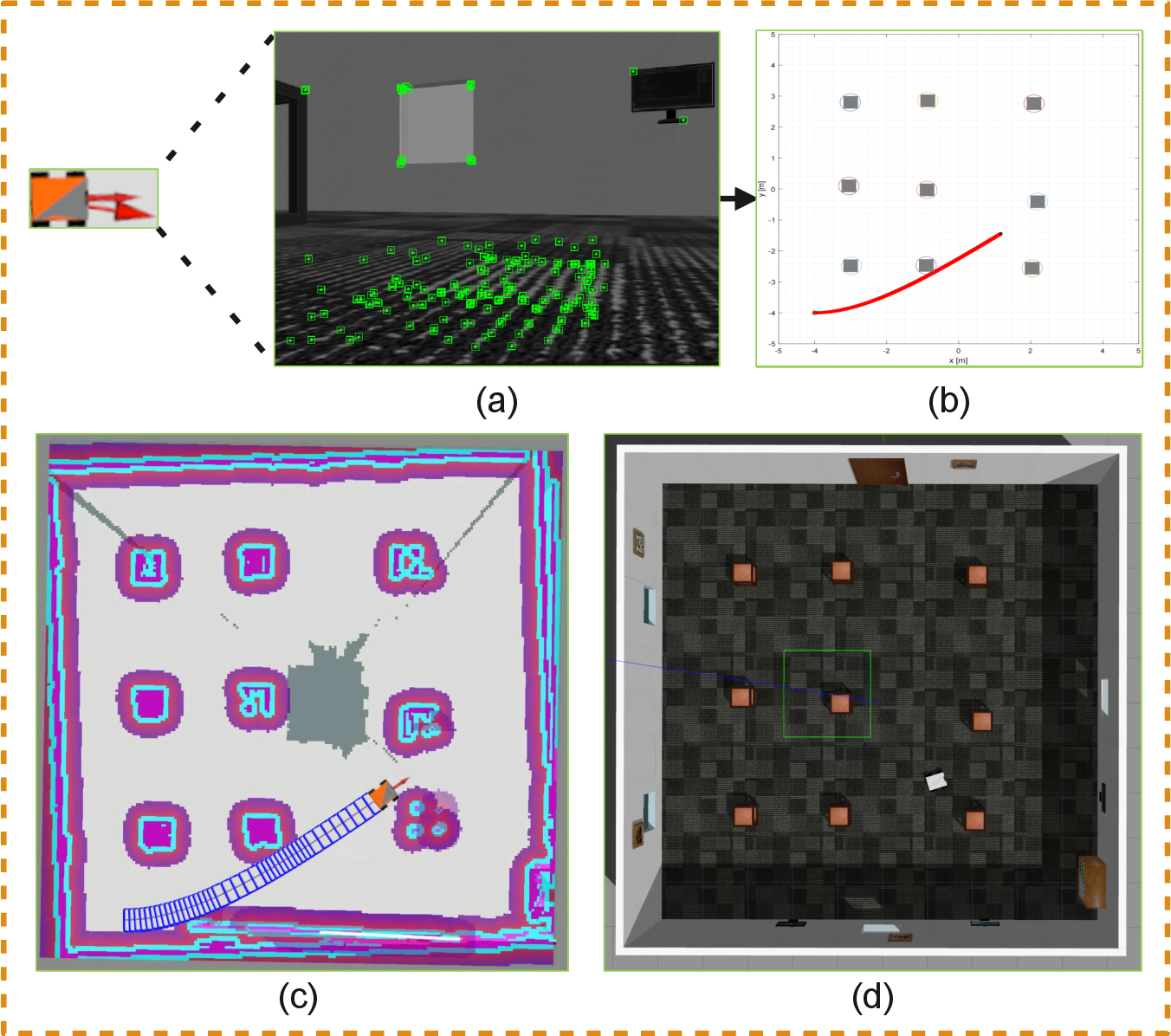}
\caption{A general depiction of our method. (a) The graph depicts the key points extracted from the scene by ORB-SLAM3. (b) The graph shows the obstacle avoidance path planning performed based on the information provided by ORB-SLAM3. (c) and (d) The graphs display the specific details of the simulated scene.}
\label{figure1}	
\end{figure}
Unmanned vehicles are often limited by heavy sensors (Radar, LiDAR, etc.), which can lead to short battery life, high costs, and large sizes, limiting the use cases and performance of these algorithms. Thus, choosing the most suitable sensors for unmanned vehicles is a critical task.

Common sensors used for these tasks include ultrasonic sensors, LiDAR, and cameras. Ultrasonic sensors excel at measuring short-range obstacles, but their accuracy decreases with distance. LiDAR sensors offer rich information, high precision, long range, and a wide field of view, but are expensive and heavy, reducing unmanned vehicles' flexibility. In contrast, cameras provide extensive scene information with low power consumption, compact size, and affordability. Therefore, developing high-precision algorithms based on camera sensors is of significant practical importance.

Many works utilize deep learning to extract information from cameras \cite{park2020vision, KIM2022116742}. These methods require high-performance computing boards with GPUs to run the algorithms, which are not suitable for lightweight small vehicle platforms. Vision-based Simultaneous Localization and Mapping (SLAM) using feature extraction methods can run on CPU-only platforms, obtaining the map information of the scene to make more precise decisions and estimating the necessary pose information for the vehicle. However, additional information is still needed to enhance the SLAM system's accuracy.

When the environment is entirely constructed by perception technologies in real-time, trajectory planning for the moving vehicle is necessary. Trajectory planning is a challenging module, as there are requirements from both the moving vehicle and the environment. From the vehicle’s perspective, the trajectory must be stable to ensure good stability and comfort. i.e., avoiding damping and sharp changes. From the environment’s perspective, there is a series of obstacles that increase the risk of collision. Therefore, a method addresses both stability and safety is needed. Control Lyapunov function (CLF) with stability constraints can be combined with the control optimization process to enhance the stability of the control system. Control barrier function (CBF) is used in RGB-D space to improve driving safety \cite{abdi2023safe}. CLF-CBF-QP is used to ensure the stability and safety of mobile car trajectory planning \cite{desai2022clf}.

Recent works predominantly utilize depth maps and poses estimated via odometry as inputs for corresponding path-planning tasks \cite{KIM2022116742, doi:10.1126/scirobotics.abg5810, 9981574, 10477312, 9309347}. These approaches often rely on IMUs, GPS, or LiDAR for pose calculation. Meanwhile, pose and depth estimation cannot be effectively globally optimized together, which fails to handle the noise introduced by depth estimation and odometry. This significantly impairs the performance of the planners. Some research employs deep learning to reduce uncertainty in depth estimation and uses reinforcement learning to deal with the noise in inputs \cite{KIM2022116742, doi:10.1126/scirobotics.abg5810}, but it places higher demands on the GPUs and CPUs onboard the vehicles. This restricts the application of such works in scenarios with limited computational resource.

This paper proposes a new autonomous obstacle avoidance algorithm for lightweight self-driving cars. We use the state-of-the-art visual SLAM algorithm ORBSLAM3 to perceive the scene and enhance the performance by eliminating outliers with optical flow epipolar constraints. ORB-SLAM3 is capable of joint optimization of pose and 3D mapping using only the CPU, and it can merge submaps to achieve real-time reconstruction of large-scale scenes whilst storing their 3D maps. Based on the rich pose and 3D information of the scene obtained, we adopt a new autonomous obstacle avoidance algorithm for lightweight self-driving cars. We employ a trajectory generation method combined with an obstacle shape reconstruction process (SRP) in irregular-shaped environments (CLF-CBF-QP-SRP) for global planning. Global planning trajectories with safety and stability can provide a general reference trajectory for self-driving car, together with a local planning using TEB (Timed Elastic Band) for local planning to ensure avoiding the local collisions. As exemplified in Fig.~\ref{figure1}, our method can effectively avoid obstacles in the scene whilst minimizing unnecessary movement.

The contributions of this work are summarized as follows:
\begin{itemize}
\item We have introduced a new lightweight single-camera-based visual SLAM system  enhanced with optical flow for outliers culling that can perceive rich information about the environment and avoid obstacles within it.
\item We have introduced a new path planning algorithm for irregular environment, which is a global planning using CLF-CBF-QP-SRP enhanced with local planning of TEB. This system can achieve collision-free path planning to various target points.
\item In addition to path planning, our approach enhances the robustness of trajectory generation by ensuring dynamic stability throughout the movement from the initial point to the target point.
\end{itemize}
\section{RELATED WORKS}

\subsection{Geometric Methods Enhanced SLAM Approach}
Visual SLAM has been improved through a series of geometric techniques. Sun \emph{et al.} \cite{SUN2018115} used additional optical flow information and a foreground model based on depth maps to eliminate outliers in the scene. Cheng \emph{et al.} \cite{doi:10.1080/01691864.2019.1610060} employed the fundamental matrix to enhance the additional information provided by the LK sparse optical flow to further increase the algorithm’s ability to outliers.

Recent work \cite{9145704} has utilized the relationships between points to filter stable data associations in the scene, and \cite{9210559} used multi-frame rather than single-frame historical observations to further identify outliers in the scene. However, these techniques do not address outliers in certain specific scenes and are usually limited to certain types of cameras. In addition, some works \cite{9981238, rs15071893, 2023MeScT..34h5202Z} have introduced a significant amount of navigation-irrelevant information and often involve lengthy processing times.

\subsection{Trajectory Generation Methods}
The process of path planning is important for moving robot cars to reach the target point safely and with good stability. A series of methods are proposed for the trajectory generation  \cite{mir2022survey}, such as rapid random tree (RRT) and Voronoi diagram-based method. RRT can be used to solve motion planning problems for robots to move from one state to another whilst avoiding obstacles by generating a space-filling tree to effectively find an optimal path. However, RRT has a major limitation that the solution may not be optimal, as the convergence rate is uncertain. Therefore, some adjusted versions of RRT are propsed, such as \cite{Wang2020LSTM}. A Voronoi diagram-based method for trajectory planning is proposed in \cite{4276103},  with simplicity, versatility and efficiency. \cite{bhattacharya2008roadmap} uses Voronoi diagram-based method combined with a roadmap to find the shortest path. \cite{9430686} uses Voronoi diagram to exclude collisions in the free space. \cite{ayawli2021path} uses Voronoi diagram to generate a safe path among the road map. However, the performance of the Voronoi diagram heavily relies on cell distribution. In areas with sparse cells, the effectiveness of Voronoi diagram diminishes. 

\begin{figure*}[t]
      \centering
      \includegraphics[width=\linewidth]{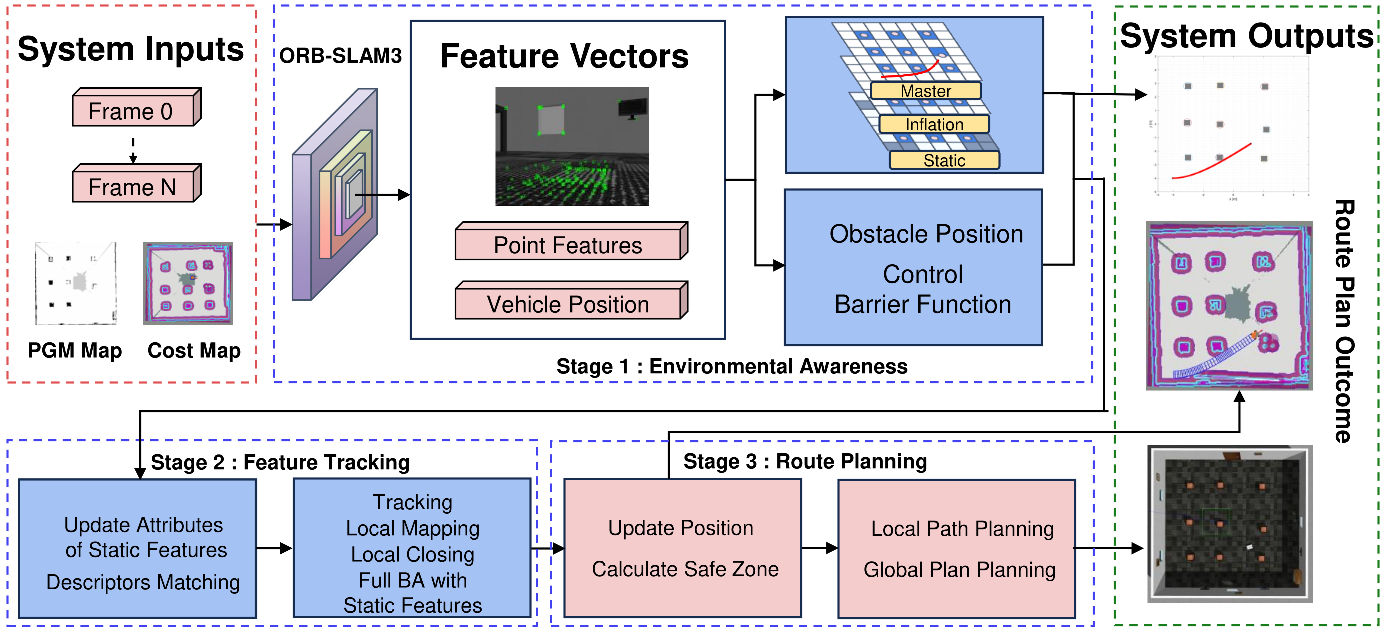}
      \caption{Our System Workflow. This system comprises two main components: environment perception and path planning. Initially, a PGM map and cost map are constructed. The vehicle, equipped with a visual sensor, extracts point features from the environment and uses relocation to ascertain its position and identify obstacles. A static map is inflated for navigational safety. The vehicle's pose is dynamically updated by tracking map points, and a global path is mapped using CBF. For local path planning, the TEB algorithm is employed. The system updates the vehicle's pose in real-time, calculates safe passage areas with CBF, and facilitates optimal, obstacle-free path selection to the destination.}
      \label{figure2}
\end{figure*}

Other techniques like the use of artificial potential filed (APF) in \cite{triharminto2016novel} for trajectory planning with multiple obstacles has also been implemented. By using the attractive and repulsive force fields on the target point and obstacles, APF guides the vehicle to the target point. A framework that combines the APF with reinforcement learning is proposed in \cite{yao2020path},  achieving collision avoidance with dense obstacles. However,  the main problem of APF is the unstable trajectory as the state of car is affected by the joint force of the attractive and repulsive force fields. In contrast to other route planning algorithms, CBF is effective for collision avoidance and enhancing safety, while CLF contributes to the stability of nonlinear systems \cite{ames2019control}. Thus, their combination promises both collision-free and stable navigation. Moreover, to address a variety of scenarios, local planning is essential to correct any discrepancies introduced by global planning. Hence, TEB \cite{rosmann2017integrated}, recognized for its proficiency in local route planning, is well-suited for this role.

\section{SYSTEM OVERVIEW}

As shown in Fig.~\ref{figure2}, our system uses visual sensors to map the environment in advance to obtain spatial boundary and obstacle information. During the process of path planning and navigation, we utilize images captured by the car's camera to generate feature points through descriptor matching. We further filter outliers in the scene using the epipolar constraints of the LK optical flow \cite{LucasKanade1981}. Finally, we use the relocation algorithm to obtain the current location and attitude information of the car. When the car is moving we use the ORB-SLAM3 algorithm to update the car's pose information in real time. After the navigation coordinate points are completely set, the system will use the pre-known static cost map for inflation. For global path planning, we use CBF to update the relative distance between obstacles and the car in real time to achieve autonomous obstacle avoidance. For complex terrain and new obstacles, we use the Timed Elastic Band (TEB) \cite{rosmann2017integrated} local planning algorithm to plan the car's path locally and use the constraints between the car and surrounding obstacles to speed up iterations to find the optimal path.

\subsection{Vehicle Model}
In this paper, to simplify the computation during the trajectory generation,  a kinematic model [26] is used. The kinematic model is governed by
\begin{equation}
\begin{split}
\dot{x} &= v\cos (\theta) \\
\dot{y} &= v\sin (\theta) \\
\dot{\theta} &= \omega
\end{split}
\end{equation}
where $v$\, denotes the velocity of the vehicle, $x$\, and $y$\, denote the longitudinal and lateral coordinates of the vehicle's midpoint, $\omega$ is the angular velocity of the vehicle, and $\theta$\, is the course angle of vehicle.  

\subsection{Perception Based on Vision} 
\begin{figure}[t]
\centering
\includegraphics[width=0.5\columnwidth]{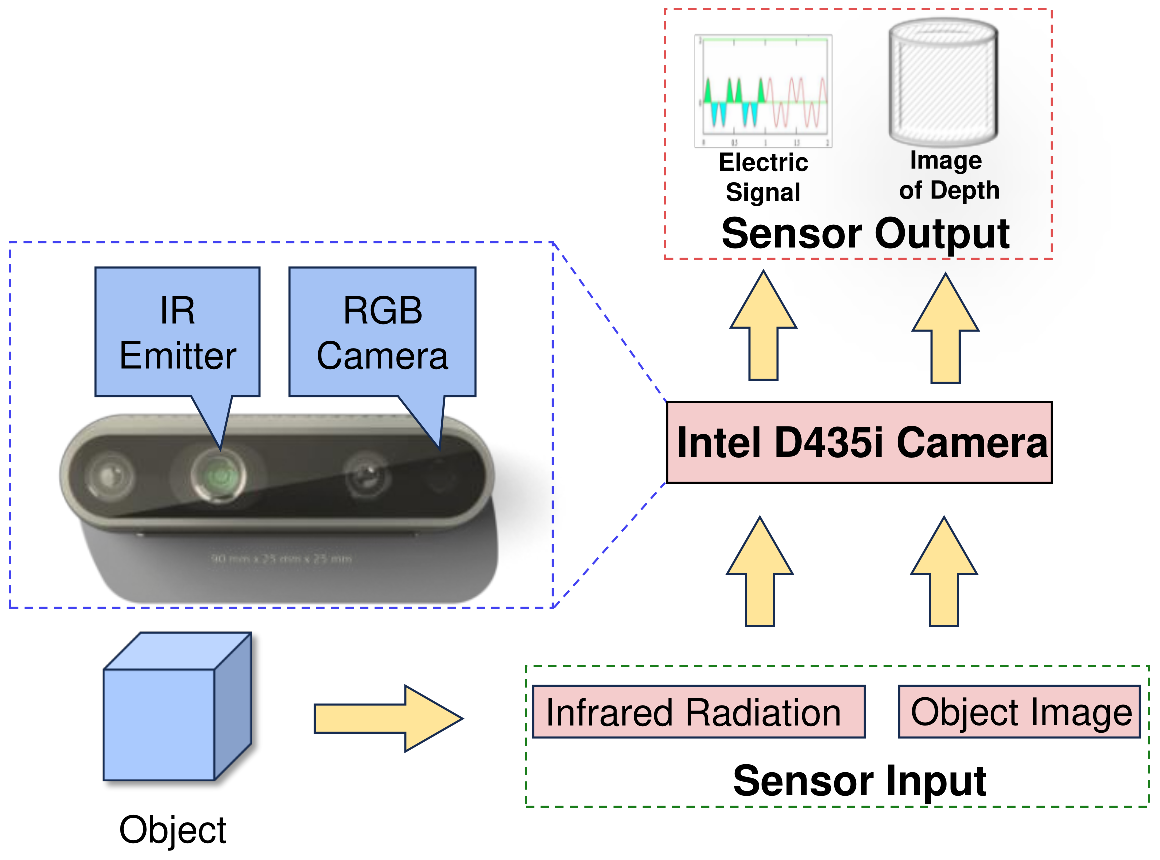}
\caption{The Intel D435i RGB-D camera utilizes the structured light triangulation method for depth sensing.}
\label{figure3}
\end{figure}

Compared with LiDAR, visual sensors have the advantages of low cost and small size and are widely used in various autonomous driving platforms. This article uses a visual SLAM system based on a RGB-D camera as shown in Fig.~\ref{figure3}, which ensures positioning accuracy and saves costs, facilitating subsequent deployment on low-cost unmanned vehicles and realizing the project as soon as possible.

\subsubsection{Point Features Matching and Attributes Updating}

\begin{figure}[t]
\centering
\includegraphics[width=0.6\columnwidth]{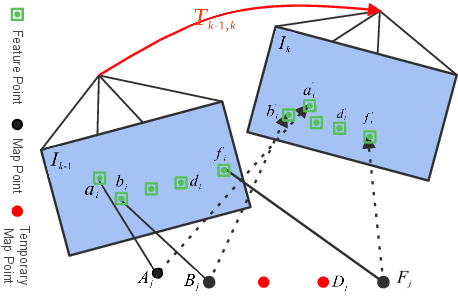}
\caption{Illustration of the point feature matching process in two frames using grid IDs, Euclidean distance, and cosine similarity to ensure alignment and temporal consistency between consecutive frames.}
\label{figure4}
\end{figure}

Our system employs a comprehensive point feature-matching methodology for stereo camera configurations. The initial step involves identifying point features from the last frame and current from the RGB-D camera that share the same ``grid ID'' of these points. Each point feature is assigned a unique ``grid~ID'' based on its spatial locality, facilitating the association of features between frames. To ensure temporal consistency, we match features by comparing descriptors and selecting the closest match based on the minimal Euclidean distance:
\begin{equation}
    \text{Distance} = \min_{j} \| \text{desc}_{\text{prev}}(i) - \text{desc}_{\text{curr}}(j) \|
\end{equation}
where \(\text{desc}_{\text{prev}}(i)\) and \(\text{desc}_{\text{curr}}(j)\) are descriptors of points from previous and current frames, respectively.
Further validation is performed by examining the cosine similarity of the angle between the direction vectors of matched points, ensuring the matched points align accurately with the expected motion model. \begin{equation}
    \text{Cosine Similarity} = \cos^{-1}\left(\frac{\vec{v}_{\text{prev}} \cdot \vec{v}_{\text{curr}}}{\|\vec{v}_{\text{prev}}\| \|\vec{v}_{\text{curr}}\|}\right)
\end{equation}
where \(\vec{v}_{\text{prev}}\) and \(\vec{v}_{\text{curr}}\) are the direction vectors of points in consecutive frames. The point features matching process among consecutive frames is visualized in Fig.~\ref{figure4}. In the figure, $T_{k-1,k} \in SE(3)$ represents the relative pose transformation between frames or the last and current image frames. Point features in 3D space, defined by a point $B_j$ and another point $F_j$, result in two points: the point $a_i$ and another point $f_i$ when projected onto the image coordinate system $I_{k-1}$ at time $t-1$. At the time $t$, the same point projected onto image coordinate system $I_k$ results in new points: point $a_i'$ and another point $f_i'$.

Bidirectional cosine similarity is used to match point pairs across frames based on grid IDs, discarding pairs below a similarity threshold. For RGB-D camera frames, we compute feature points' grayscale centroids and direction vectors, which are crucial for pose optimization during bundle adjustment. 

\subsection{Outlier Features Removing}
In our SLAM system, we mitigate the impact of inconsistent feature points on pose estimation by implementing the Lucas-Kanade (LK) method to enforce epipolar constraints, guided by the fundamental matrix \( F \). This ensures geometrically coherent feature point pairs by setting a threshold distance from the epipolar line, allowing us to filter out mismatches. This fusion of techniques bolsters system stability.

To refine feature selection, we eschew traditional Harris corner ~\cite{HarrisStephens1988} matches in favor of those with lower disparity in central pixel blocks. We then apply a stringent distance metric to eliminate outliers, which is central to the precision of our algorithm.

For rigor, the RANSAC algorithm~\cite{FischlerBolles1981} is employed to extract the fundamental matrix \( F \) that maximizes inlier correspondences. With this matrix, we map features from one frame to their epipolar counterparts in the subsequent frame.

Consider matched points \( p_1 \) and \( p_2 \) in consecutive frames, with homogeneous coordinates \( \mathbf{P}_1 = [u_1, v_1, 1]^\top \) and \( \mathbf{P}_2 = [u_2, v_2, 1]^\top \), respectively. The epipolar line \( \mathbf{L}_1 \) for \( \mathbf{P}_1 \) is obtained as \( \mathbf{L}_1 = F \mathbf{P}_1 \). We define the distance to the epipolar line as
\begin{equation}
D = \frac{|\mathbf{P}_2^\top F \mathbf{P}_1|}{\sqrt{F_{row1}^2 + F_{row2}^2}}
\end{equation}
where \( F \), \(F_{row1}\), and \(F_{row1}\) represent the fundamental matrix, the first and second rows of the fundamental matrix \( F \), respectively. Points with \( D \) exceeding a predefined threshold are deemed outliers and discarded.

\subsection{Point Features Optimizing Algorithm}
In 3D visual SLAM systems, map point features play a critical role in environment modelling and localization. However, factors like sensor noise and dynamic environments can induce errors in the orientation and position of these features. To address this, we adopt an optimization-based reference keyframe pose correction method to enhance the accuracy and consistency of map point features.

The process of pose correction is detailed as follows: Let $R_{wr}$ and $t_{wr}$ represent the rotation matrix and the translation vector from the reference keyframe to the world coordinate system, respectively. We assume the world coordinates of the map point features to be corrected as $P_{3Dw}$, with $P_{3Dw_{sp}}$ denoting the coordinate of the point feature. By using the Sim3 transformation matrices $S_{rw}$ and $corSwr$, we transform the pose of map point features from the reference keyframe of the current frame to the corrected reference keyframe as follows:
\begin{equation}
CorP_{3D_{w_{sp}}} = corSwr \times (S_{rw} \times P_{3D_{w_{sp}}})
\end{equation}
where $\times$ denotes matrix multiplication, and $CorP_{3Dw_{sp}}$  represents the coordinates of the corrected map point feature in the corrected reference keyframe coordinate system. 
Enhancing the precision of these transformations directly impacts the accuracy of the camera pose and the overall SLAM performance. By refining the keyframe poses through these corrections, we effectively reduce the impact of positioning errors caused by sensor noise and dynamic environmental factors. This sets a robust foundation for subsequent optimization processes such as motion-only bundle adjustment (BA).

Motion-only BA optimizes the camera orientation \( R \in SO(3) \) and position \( t \in \mathbb{R}^3 \), by minimizing the reprojection error between matched 3D points \( X'_i \) in world coordinates and their corresponding image keypoints \( x'_i \), which may be either monocular \( x_m^i \in \mathbb{R}^2 \) or stereo \( x_s^i \in \mathbb{R}^3 \),  with \( i \in \Lambda \) the set of all matches, as follows:
\begin{equation}
    \{R, t\} = \arg\min_{R,t} \sum_{i \in \Lambda} \rho  \left\| x'_i - \pi_d \left( R X'_i + t \right) \right\|_{\Sigma}^2 
\end{equation}
where \(\rho\) is the robust Huber cost function, \(\Sigma\) represents the covariance matrix related to the scale of the keypoint, and \(\pi_d\) is the projection function for RGB-D cameras defined as
\begin{equation}
\pi_d\left(\begin{bmatrix} X \\ Y \\ Z \end{bmatrix}\right) = \begin{bmatrix} f_x \frac{X}{Z} + c_x \\ f_y \frac{Y}{Z} + c_y \\ Z \end{bmatrix}
\end{equation}
where \([X, Y, Z]^\top\) represent the coordinates of a point in the world coordinate system. \(\frac{X}{Z}\) and \(\frac{Y}{Z}\) are the normalized image plane coordinates. \(f_x\) and \(f_y\) are the focal lengths of the camera along the X and Y axes, respectively. \(c_x\) and \(c_y\) are the coordinates of the principal point, typically at the center of the image. \(Z\) is the depth value directly measured by the RGB-D sensor, providing real-time depth at each image pixel.

Local BA optimizes a subset of covisible keyframes \(K_L\) and all points observed in those frames \(P_L\). Non-optimized keyframes \(K_F\), while fixed during optimization, contribute observations of \(P_L\), adding constraints to enhance map stability without altering their poses. Define \(\lambda_k\) as the set of matches between points in \(P_L\) and keypoints in keyframe \(k\), the optimization is formulated as follows:
\begin{equation}
    \{\mathbf{X}_i, R_l, t_l \mid i \in P_L, l \in K_L\} = \arg\min_{\mathbf{X}_i, R_l, t_l} \sum_{k \in K_L \cup K_F} \sum_{j \in \lambda_k} \rho ( E_{kj} )
\end{equation}
where \(\mathbf{X}_i\) represents the 3D position of the \(i\)-th point in \(P_L\). \(R_l\) and \(t_l\) are the rotation matrix and translation vector for the \(l\)-th keyframe in \(K_L\). Reprojection error \(E_{kj}\) is quantified as
\begin{equation}
    E_{kj} = \left\| \mathbf{x}'_j - \pi_d \left( R_k \mathbf{X}_i + t_k \right) \right\|_{\Sigma}^2
\end{equation} where \(\mathbf{x}'_j\) represents the 2D projection coordinates of the j-th feature point on the image plane. In contrast to Local BA, Full BA adjusts all keyframes and map points, except the origin keyframe which remains fixed to resolve scale ambiguity. This extensive optimization ensures the highest accuracy by refining camera poses and landmark positions across the entire map based on all available visual information.

\subsection{Global Path Planning by Using CLF-CBF-QP-SRP}
In this section, we introduce two key concepts for designing a safe and stable control system: control Lyapunov function (CLF) and control barrier function (CBF). A CLF is a positive definite function that decreases along the trajectories of the system, and can be used to ensure asymptotic stability of a desired equilibrium point. A CBF is a function that satisfies some conditions on its Lie derivatives, and can be used to enforce state constraints in the operating space. By combining CLF and CBF, we can design a control law that guarantees both safety and stability of the car.

\subsubsection{The formulation of CLF}
To generate a stable trajectory, for a moving car, the vehicle model is written as the following nonlinear control system
\begin{equation}
\dot{s} =f(s)+g(s)u
\label{eq10}
\end{equation}
where $s \in \mathbb{R}^{n}$ is the state of the moving car, and $u \in \mathbb{R}^{m}$ is the control input. Functions $f$ and $g$ are smooth vector fields. The control input is subject to the following constraints:
\begin{equation}
u \in \mathcal{U} \subset \mathbb{R}^{m} := \{ u \,|\, u_{\min} \leq u \leq u_{\max} \}
\end{equation}
where $\mathcal{U}$\, is the admissible input set of the control system. $u_{\min}$\, and $u_{\max}$\, are the minimum value and maximum value of inputs. Definition 1 combines an affine constraint with $u$ to achieve an optimization-based controller.

\emph{Definition 1: }
Assume $V$\, is a Lyapunov function when the following condition is satisfied \cite{ames2019control}:
\begin{equation}
\underset{u\in \mathcal{U}}{\inf} \left[ L_{f}V(s)+L_{g}V(s)u \right]\le \textrm{K(V(s))}
\end{equation}
where $L_{f}V(s)$\, and $L_{g}V(s)$\, are the Lie-derivatives of $V(x)$\,, $\textrm{K()}$\, is a class $\mathcal{K}$\, function. The class $\mathcal{K}$\, function is a function $k:(0,p] \to (0,\infty]$\, with the property of strictly increasing and the initial value $k(0)=0$\,. Then $s$\, can be stabilized by the following equation
\begin{equation}
K_{CLF}(s):=\left \{ u\in \mathcal{U}, L_{f}V(s)+L_{g}V(s)u\le \textrm{K(V(s))}\right \}.
\label{eq13}
\end{equation}

\subsubsection{The stability control of moving car using CLF}
 As the basic formulation CLF is introduced in ~\eqref{eq13}, the problem is to combine CLF with the dynamics of moving car. Assume the current position of moving car is $p_{c}=(x_{c}, y_{c}, \theta_{c})$\, and the target position is $p_{t}=(x_{t}, y_{t}, \theta_{t})$\,. The error between the current position and the final position, $e=[x_{c}-x_{t}, y_{c}-y_{t}, \theta_{c}-\theta_{t}]$\,, can be used for building the CLF as follows
\begin{equation}
V(s)=ePe^{T} 
\end{equation}
where $P$\, is a 3x3 symmetric matrix with five parameters used to ensure that the $V(s)$ is positive definite. The format of $P$\, in this paper is formulated as
\begin{equation}
P=\begin{bmatrix}  p_{1}& 0 &p_{2} \\  0& p_{3} & p_{4}\\  p_{2}& p_{4} &p_{5}\end{bmatrix}. 
\end{equation}
Therefore, \eqref{eq10} can be transferred to the format suitable for the target of moving car.
\subsubsection{The formulation of CBF}
CBF is used for the collision-free control, together with CLF for stability. A set $\mathcal{D}$ is defined as composing a continuously differentiable function $h$:
$\mathcal{Z}\subset\mathbb{R}^n\to\mathbb{R}$, yielding:
\begin{equation}
\begin{aligned}
\mathcal{D}& =\{\mathbf{s}\in\mathcal{Z}\subset\mathbb{R}^n:h(\mathbf{s})\geq0\},  \\
\partial\mathcal{D}& =\{\mathbf{s}\in\mathcal{Z}\subset\mathbb{R}^n:h(\mathbf{s})=0\},  \\
\operatorname{Int}(\mathcal{D})& =\{\mathbf{s}\in\mathcal{Z}\subset\mathbb{R}^n:h(\mathbf{s})>0\}, 
\end{aligned}
\end{equation}
where $\mathcal{D}$ is the set that achieves collision-free.

\emph{Definition 2: } The nonlinear control system is safe if $\mathcal{D}$ is forward invariant. Assume $T$\, is the time interval, the set $\mathcal{D}$ is forward invariant when each
$\mathbf{s}_0\in\mathcal{D},\mathbf{x}(nT)\in\mathcal{D}\text{ for }\mathbf{s}(0)=\mathbf{x}_0,\forall n\geq0$ \cite{huang2023obstacle}.

\emph{Definition 3: }The function $h$ is a CBF defined on the set $\mathcal{Z}$, if there exists an extended class $\mathcal{K}_\infty$ function $\alpha$ such that the nonlinear control system satisfies \cite{ames2019control}:
\begin{equation}
  \sup_{u\in\mathcal{U}}[L_fh(s)+L_gh(s)u]\geq-\alpha(h(x))
  \label{eq17}
\end{equation}
where $L_fh(s)$ and $L_gh(s)$ are Lie-derivatives of $h(s)$.

The set of controls that allow $\mathcal{D}$ collision-free
for all $s\in\mathcal{Z}$ are formulated as 
\begin{equation}\label{eq18}
  K_\mathrm{cbf}(s):=\{u\in\mathcal{U},L_fh(s)+L_gh(s)u\geq-\alpha(h(s))\}.
\end{equation}

\subsubsection{Safe control of moving car using CBF}

\begin{figure}[t]
\centering
\includegraphics[width=0.3\columnwidth]{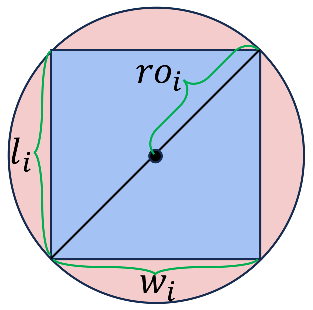}
\caption{Illustration of SRP for an obstacle.}
\label{figurelabel5}
\end{figure}

In this paper, all obstacles are considered to be static with square shapes, in contrast to the regular circular shapes used for CBF calculations. To compute CBFs, the SRP (Specific Required Parameter) of each obstacle must be determined. As illustrated in Fig.~\ref{figurelabel5}, a circle is used to enclose the $i^{th}$ obstacle, whose radius is calculated as
\begin{equation}\label{eqn10}
  r_{i}=\sqrt{ \left( \frac{l_{i}}{2} \right)^{2} + \left( \frac{w_{i}}{2} \right)^{2}} 
\end{equation}
where $l_{i}$ and $w_{i}$ are the length and width of the $i^{th}$\, obstacle. To reach the target point in an open space, acquired by the perception stage, there is a moving car together with a set of $\mathbb{N}$ obstacles. Each obstacle is represented by
$\mathrm{O}_i\in\mathbb{O}=\{\mathrm{O}_0,\mathrm{O}_1,\dots,\mathrm{O}_{N-1}\}$. 
Assume the position of the
obstacle $\mathrm{O}_i$ is denoted by $\mathbf{z}_{\mathrm{O}_i} = (x_{\mathrm{O}_i},y_{\mathrm{O}_i})$, thus \eqref{eq17} is converted to
\begin{equation}\label{eqn14}
\sup_{u\in\mathcal{U}} \left[ L_fh_i(s)+L_gh_i(s)u+\frac{\partial h_i(s)}{\partial t} \right] \geq -\alpha(h_i(s)).
\end{equation}
Thus, for $\mathbf{u}\in\mathcal{U}$\,, the set of controls that ensures the robot car to be safe is expressed as:
\begin{equation}
K_{\mathrm{cbf}}^i(s):= \left\{L_fh_i(s)+L_gh_i(s)u \ge-\alpha(h_i(s)) \right\}.
\label{eq21}
\end{equation}

Assuming that the robot car is defined with a safe radius of $r_s$, the safe distance between the car and obstacle $\mathrm{O}_i$ is defined as $r_i = r_s + r_{\mathrm{O}_i}$. Then the CBF is designed as
\begin{equation}\label{eqn16}
  h_i(s)=(x_\text{c}-x_{\mathrm{O}_i})^2+(y_\text{c}-y_{\mathrm{O}_i})^2-r_i^2.
\end{equation}

\subsection{Safe and Stable Control for Self-driving Cars}
Since both the constraints of accurate and safe control have the affine form, real-time solutions can be acquired. Therefore, a QP-based controller that combines CBFs for safety and CLF with \eqref{eq13} and \eqref{eq21} for stability is as follows:\\
\textbf{CLF-CBF-QP:}
\begin{subequations}
\begin{align}
& \hspace{1.5em} \min_{(u,\delta)\in\mathbb{R}^{m+1}} \frac{1}{2}u^{T}Hu+p\delta^{2}+(u-u_{l})^{T}Q(u-u_{l}) \label{eq23a}  \\
\mathbf{s.t. \,\,}  
\quad & L_{f}V(s)+L_{g}V(s)u+K(V(s))\leq\delta  \\
&L_{f}h_{i}(s)+L_{g}h_{i}(s)u+\alpha(h_i(s)) \geq0, ~i=0,1,\ldots,N-1 \\
& u\in\mathcal{U}
\end{align}
\end{subequations}
where the objective function \eqref{eq23a} is divided into three parts: the first part minimizes the magnitude of $\mathbf{u}$, the second part adds an extra quadratic cost, and the third part ensures the smoothness of $u$. $H$ and $Q$ are positive definite matrices, $p > 0$ is the weight coefficient of the relaxation variable $\delta$\, and $u_{l}$ is
the control value of the last moment.

\section{EXPERIMENTAL EVALUATION}


The simulations were conducted to verify the safety, stability and efficiency of the proposed route planning algorithm. The experiments were conducted on a Linux machine with the Ubuntu 18.04.6 LTS OS, a 12th generation 16-thread Intel\textsuperscript{\textregistered}Core\texttrademark\ i5-12600KF CPU, an NVIDIA GeForce RTX 3070Ti GPU, and 16GB of RAM.  The QP problem is solved using the quadprog solver in MATLAB R2022B. 

In order to verify the effectiveness of our proposed system, we built a closed square experimental scene. In order to ensure that the visual algorithm can extract a certain number of feature points in the surrounding environment, we added TVs, murals and other items to the inside of the room walls. Among them, we use 9 square hollow tables as obstacles. The feature points in the middle of such hollow objects are often not on the obstacles but on the wall behind them. Such a scenario poses certain challenges to the perception algorithm to test. The car is placed on the map as the carrier of the system, rather than the car's preset starting point. This can effectively test the effectiveness of our system's relocation system. The car in the experiment is rectangular in shape and has four wheels, equipped with an RGB-D vision sensor on the top of the front end. The initial verification of the obstacle avoidance algorithm(CLF-CBF-QP) will be verified in Matlab, and the experiments of the whole visual navigation system are carried out in simulation environments Gazebo and Rviz. Fig.~\ref{figurelabel6} illustrates one experiment on how to navigate the car from the start point to the destination. 


\subsection{Simulation Environments Setup}

\begin{figure}[t]
\centering
\includegraphics[width=0.7\columnwidth]{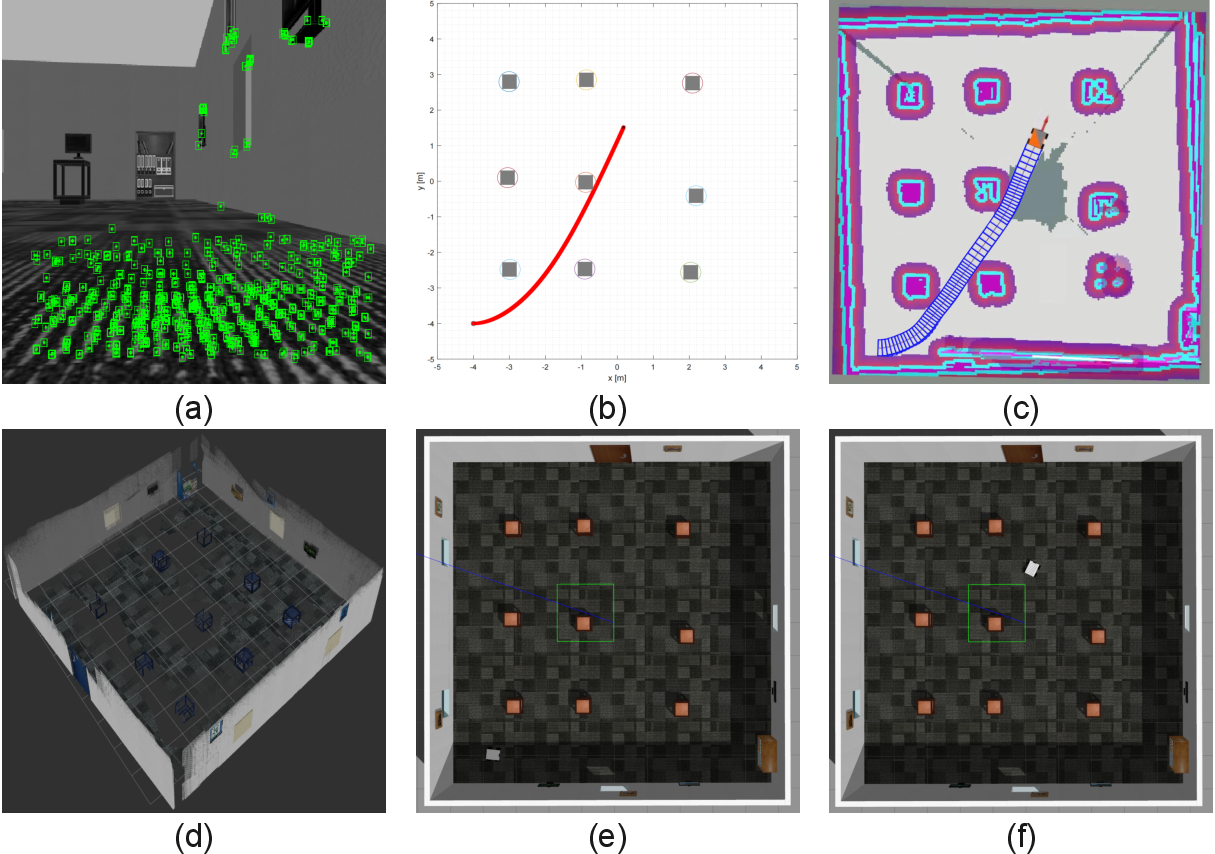}
\caption{Illustrative experiment on how to navigate a robot car from the start point to the destination.}
\label{figurelabel6}
\end{figure}

\begin{figure}[ht]
\centering
\includegraphics[width=0.3\columnwidth]{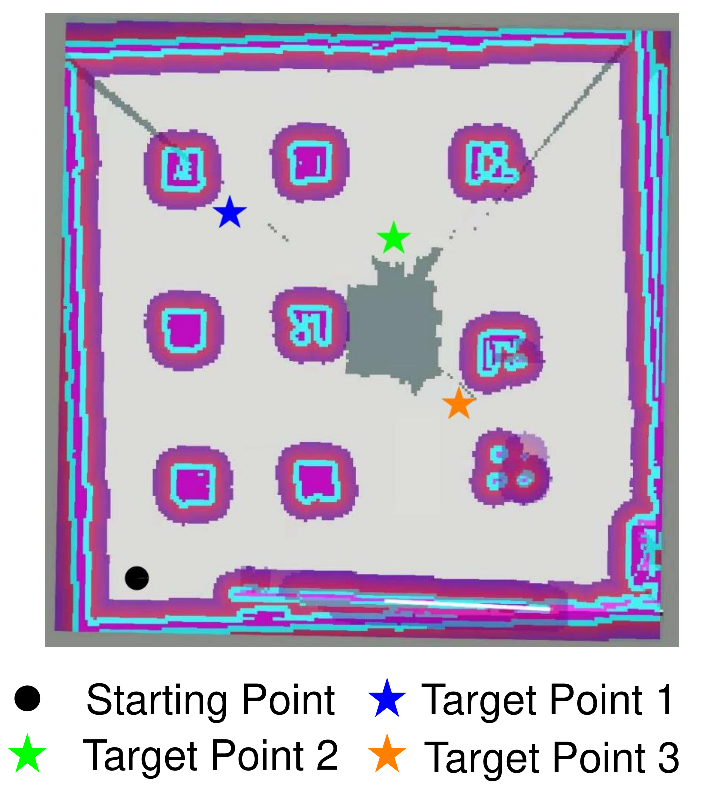}
\caption{The three different destinations chosen in the experiment.}
\label{figurelabel7}
\end{figure}

We place the car at the starting point with coordinates (-4, -4) and the target points with coordinates (0, 1.5). First, the car uses the visual sensor to detect the feature points in the image, as shown in Fig.~\ref{figurelabel6}(a). After comparing it with the pre-saved atlas, it uses the relocation algorithm to obtain the current location of the car, and updates the car's posture and obstacle distance. Then, the velocity and direction of the car are updated by the route planning algorithm. In order to verify the effectiveness of our proposed algorithm, we first performed CBF on MATLAB for motion trajectory planning test. The simulation trajectory on MATLAB is shown in the red trajectory in Fig.~\ref{figurelabel6}(b). Our system ensures that it does not encounter obstacles and selects the shortest distance between two points. In order to further verify the effectiveness of our algorithm, we used the ROS system to conduct further simulation tests under Ubuntu. The planned path simulation of the car is displayed as a blue box in Rviz, where the point in the middle of the blue box represents the trajectory of the vehicle. The blue box represents the space the car takes up in the environment. The colored map represents costmap, which is used to represent obstacles and passable areas in the environment. The costmap is represented as a two-dimensional grid, with each grid cell (or pixel) colored according to the ``cost'' or ``safty'' it represents. The meaning of the colors is as follows: Blue usually represents low-cost areas, i.e. areas that are relatively safe and free for the robot car. Red usually indicates very high costs and is often directly associated with obstacles. This is an area that robot cars should avoid. Grey represents unknown areas, i.e. those areas that the robot car has not yet explored or whose properties cannot be determined. White represents known free areas, i.e. areas without obstructions.

\begin{figure*}[ht]
      \centering
      \includegraphics[width=1\columnwidth]{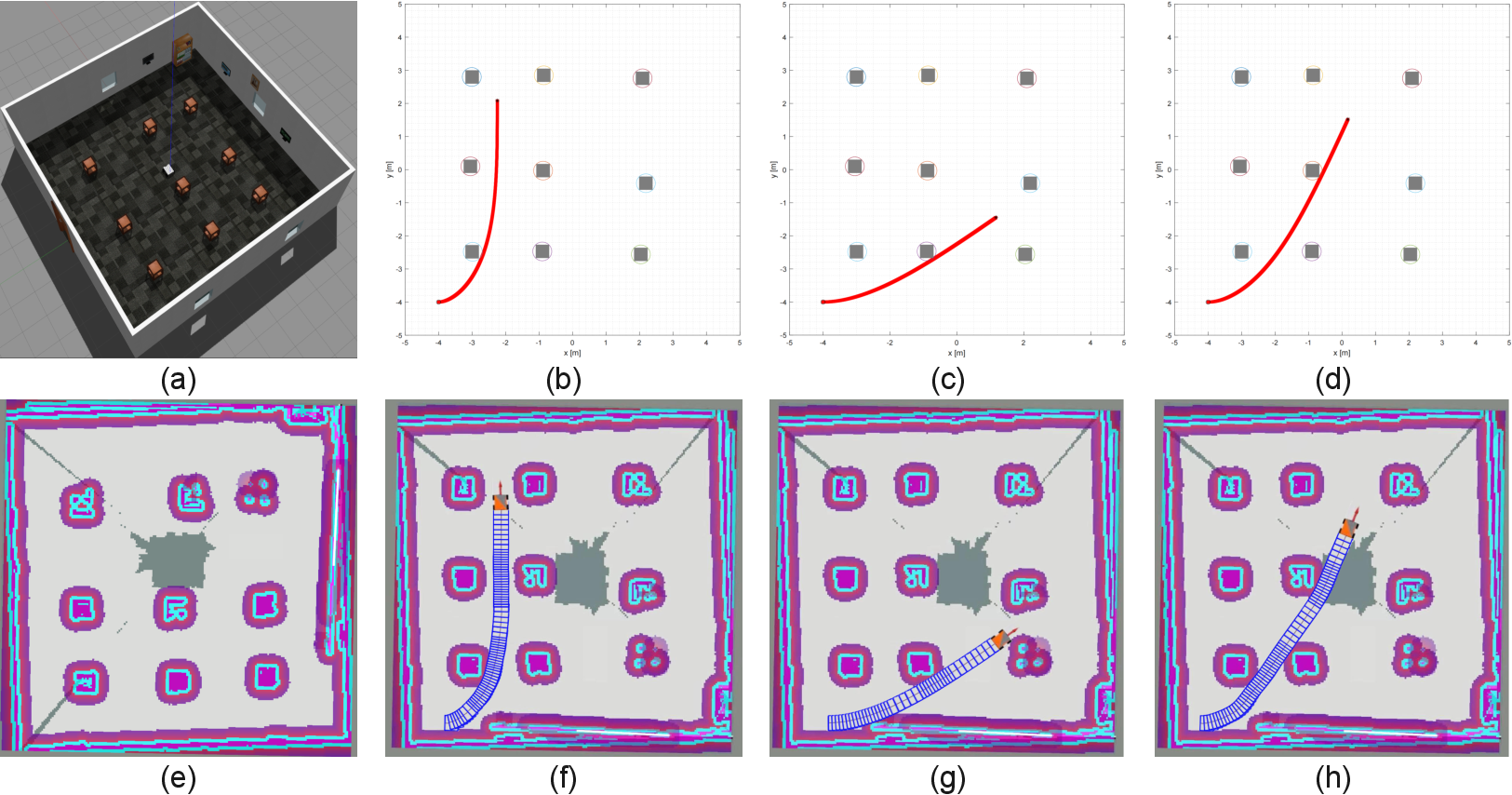}
      \caption{Experimental results for three different destinations.}
      \label{figurelabel8}
\end{figure*}

\begin{figure}[h]
    \centering
    \begin{subfigure}[b]{0.49\columnwidth}
        \includegraphics[width=0.9\textwidth]{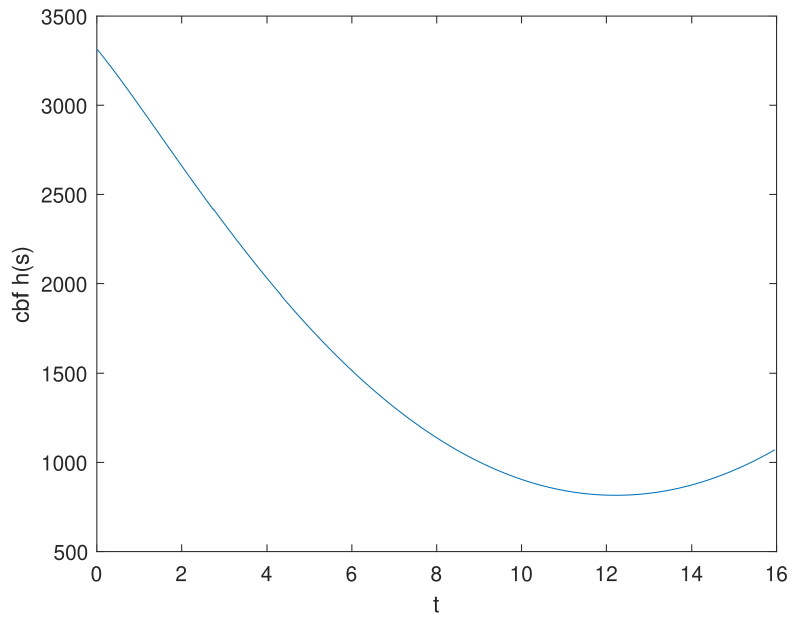}
        \caption{Variation of CBF}
         \vspace{0.2cm}
        \label{fig:1pdf}
    \end{subfigure}
    \begin{subfigure}[b]{0.49\columnwidth}
        \includegraphics[width=0.9\textwidth]{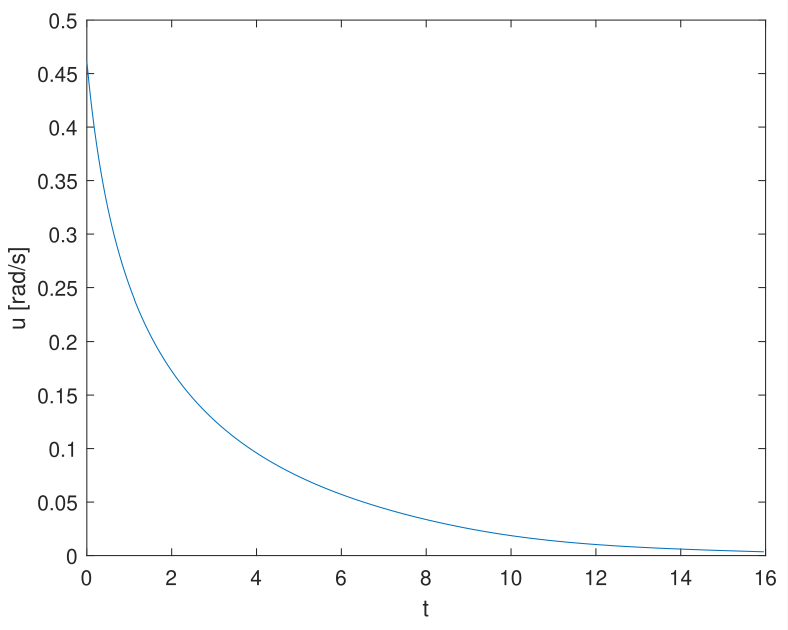}
        \caption{Variation of $\omega$}
         \vspace{0.2cm}
        \label{fig:2pdf}
    \end{subfigure}
    \begin{subfigure}[b]{0.9\columnwidth}
    \vspace{0.5cm}
    \includegraphics[width=0.8\columnwidth]{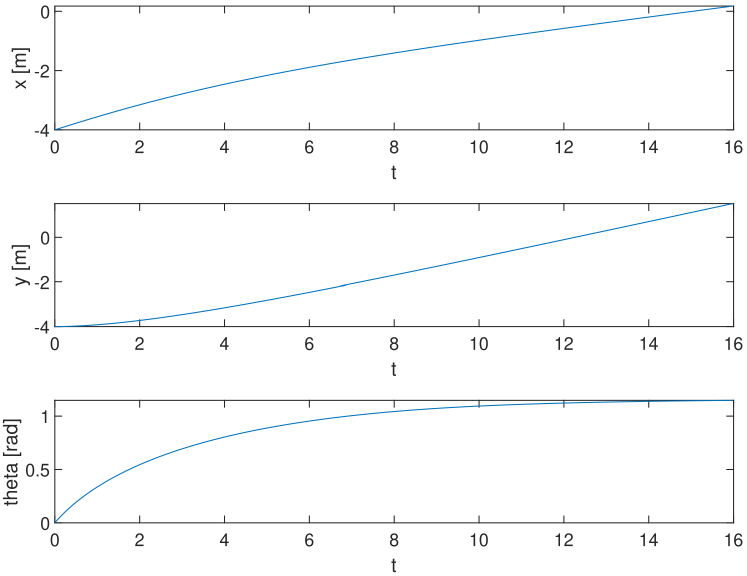}
    \caption{Variation of Longitudinal Distance, Lateral Distance and $\theta$.}
     \vspace{0.2cm}
    \label{fig:3pdf}
    \end{subfigure}
     \caption{Variation of variables for target point 3.}
    \label{fig:4pdf}
\label{figurelabel9}
\end{figure}


\begin{figure}[!ht]
    \centering
    \begin{subfigure}[b]{0.3\textwidth}
        \includegraphics[width=\textwidth]{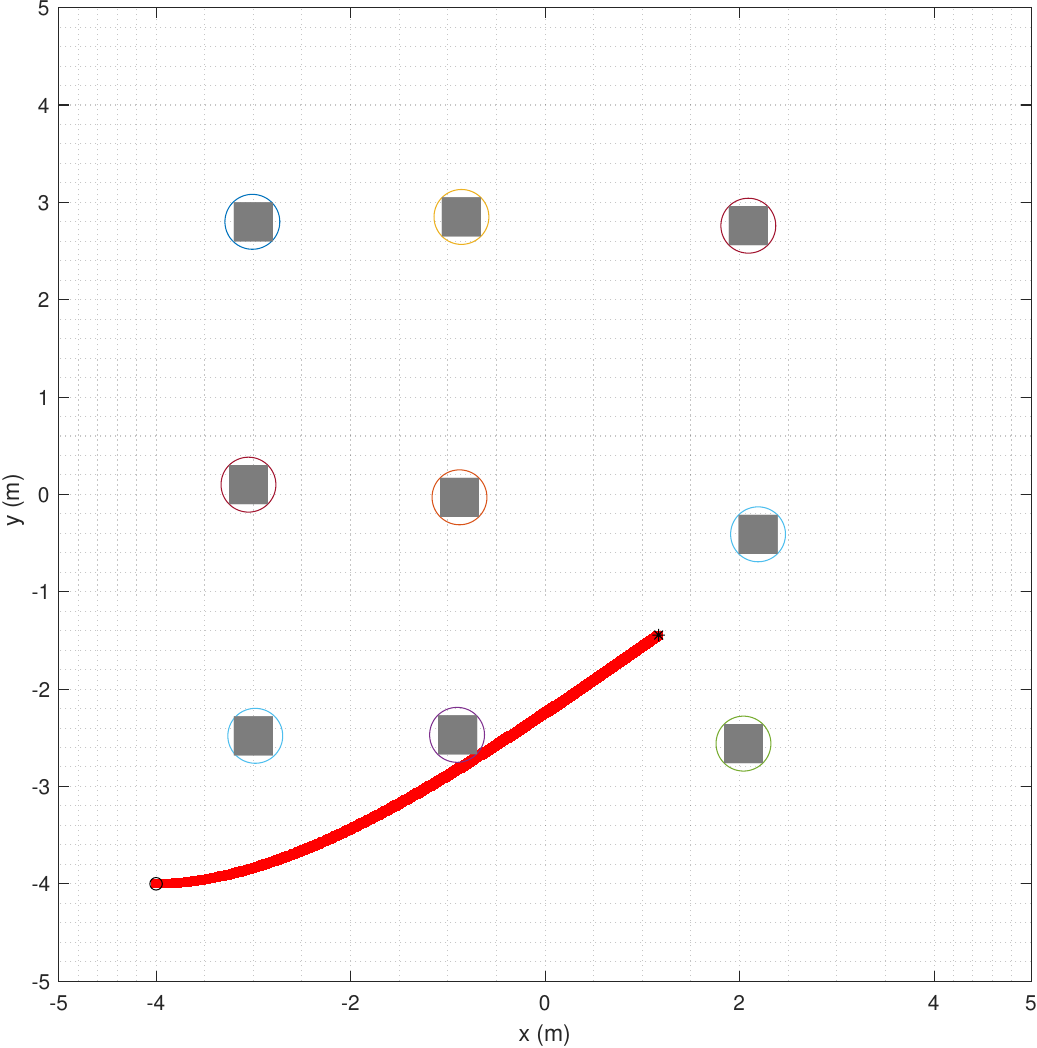}
        \caption{CLF-CBF-QP-SRP for target point 1}
        \vspace{0.2cm}
        \label{fig:1pdf}
    \end{subfigure}
    \hfill
    \raisebox{0.3cm}{
        \begin{subfigure}[b]{0.3\textwidth}
            \includegraphics[width=\textwidth]{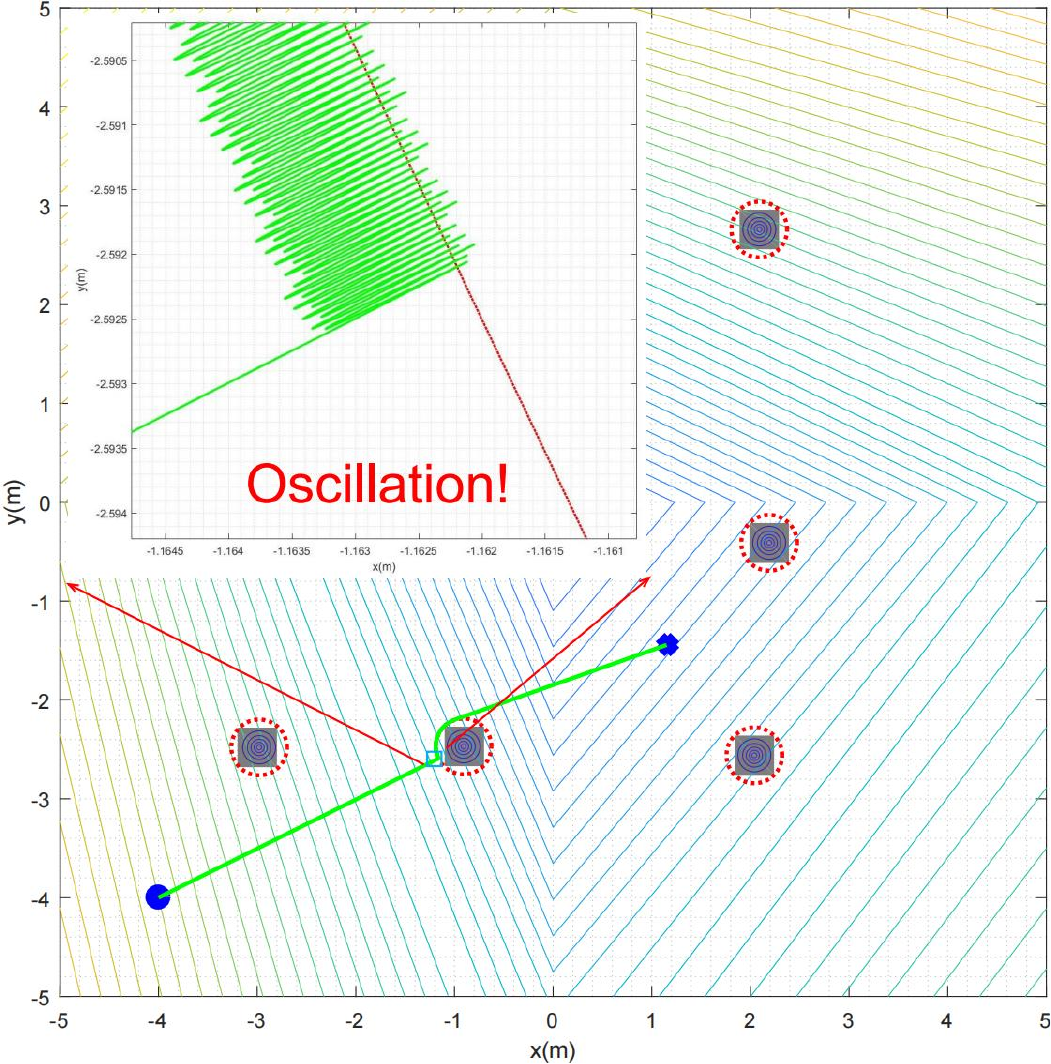}
            \caption{APF for target point 1}
            \vspace{0.3cm}
            \label{fig:2jpg}
        \end{subfigure}
    }
    \hfill
    \begin{subfigure}[b]{0.3\textwidth}
        \includegraphics[width=\textwidth]{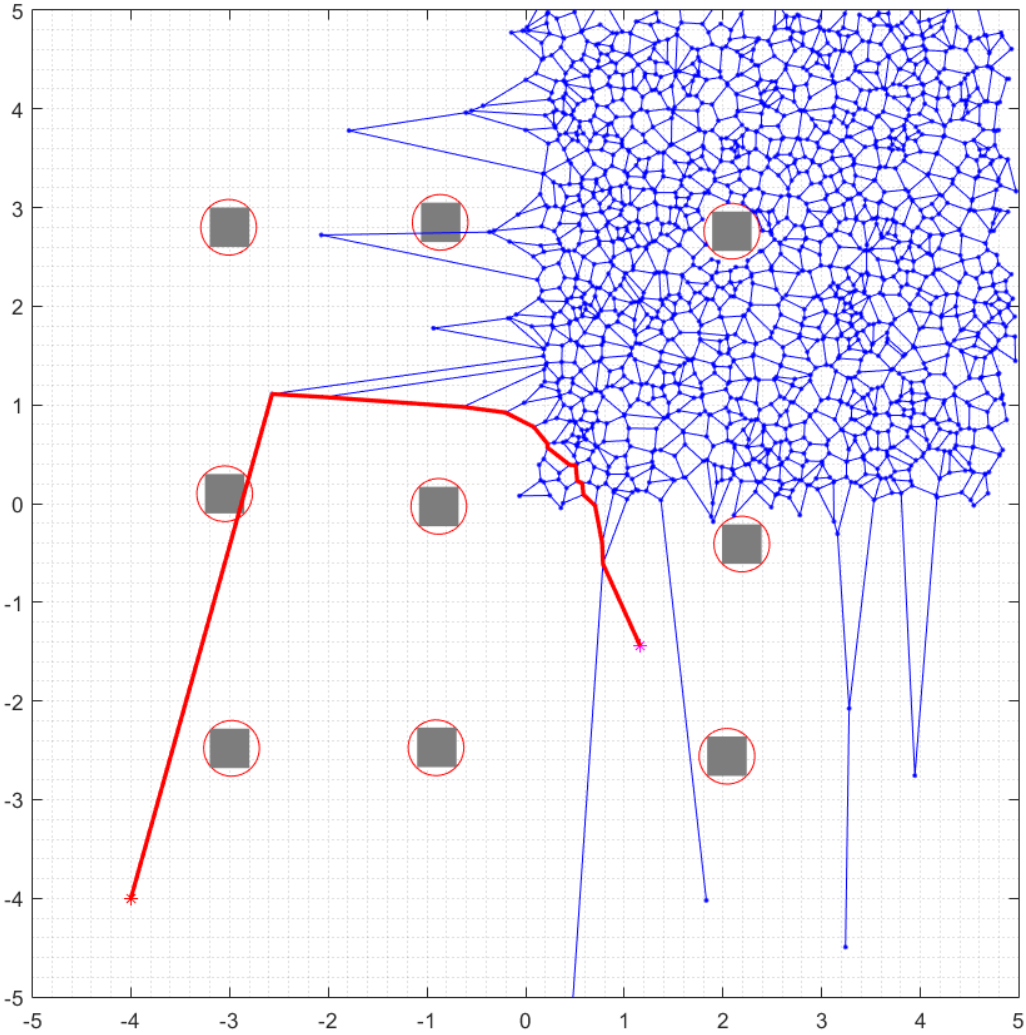}
        \caption{Voronoi diagram for target point 1}
        \vspace{0.2cm}
        \label{fig:3pdf}
    \end{subfigure}

    \begin{subfigure}[b]{0.3\textwidth}
        \vspace{0.5cm}
        \includegraphics[width=\textwidth]{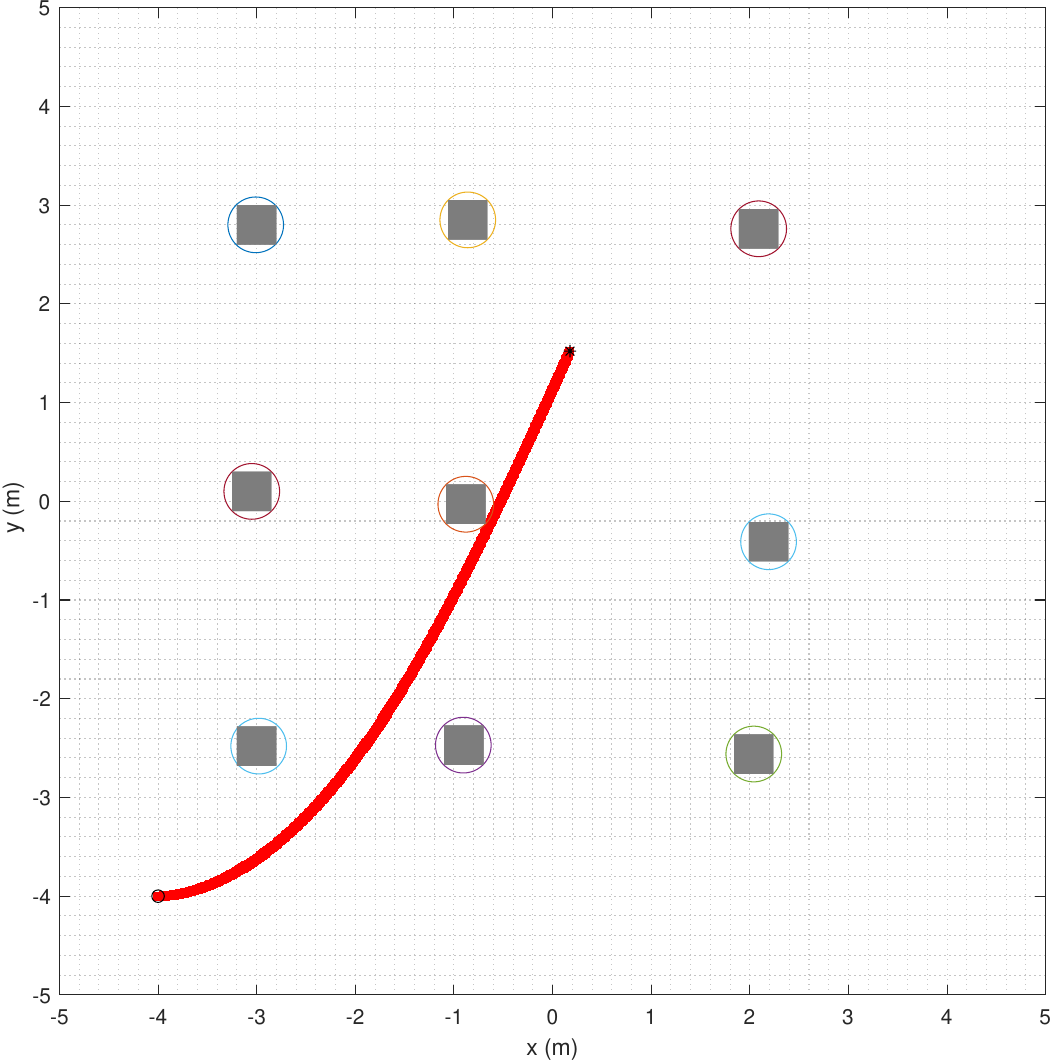}
        \caption{CLF-CBF-QP-SRP for target point 2}
        \vspace{0.2cm}
        \label{fig:4pdf}
    \end{subfigure}
    \hfill
    \raisebox{0.3cm}{
        \begin{subfigure}[b]{0.3\textwidth}
            \includegraphics[width=\textwidth]{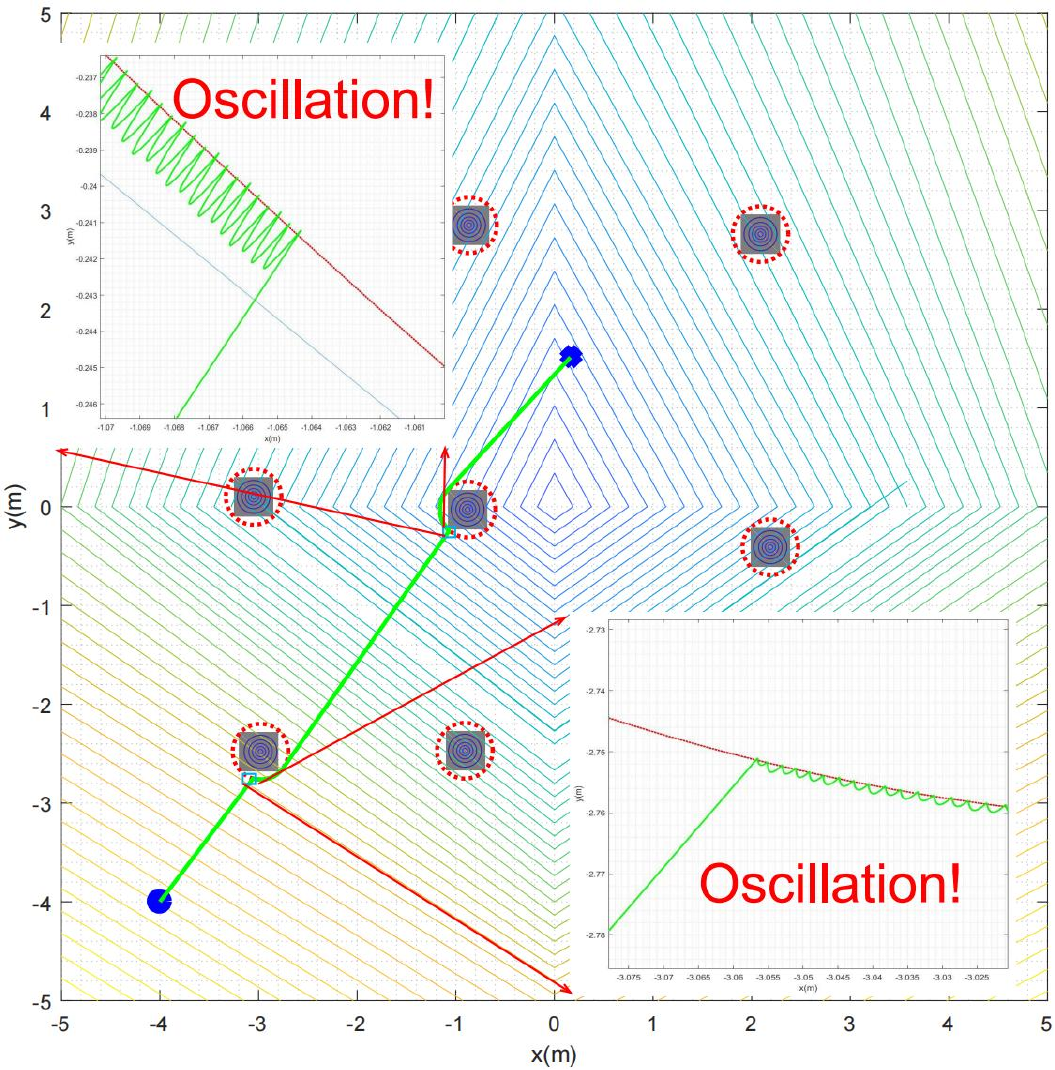}
            \caption{APF for target point 2}
            \vspace{0.3cm}
            \label{fig:5jpg}
        \end{subfigure}
    }
    \hfill
    \begin{subfigure}[b]{0.3\textwidth}
        \includegraphics[width=\textwidth]{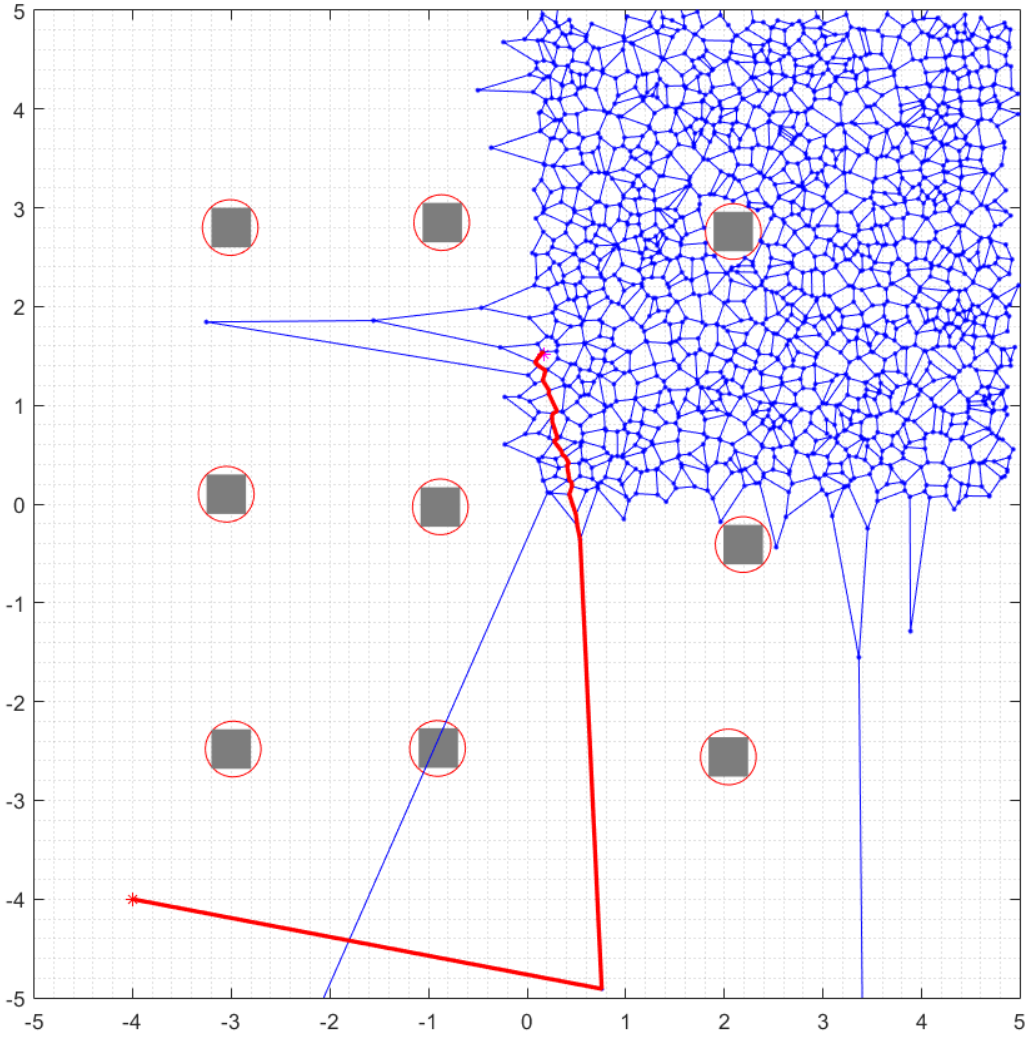}
        \caption{Voronoi diagram for target point 2}
        \vspace{0.2cm}
        \label{fig:6pdf}
    \end{subfigure}

    \begin{subfigure}[b]{0.3\textwidth}
    \vspace{0.5cm}
        \includegraphics[width=\textwidth]{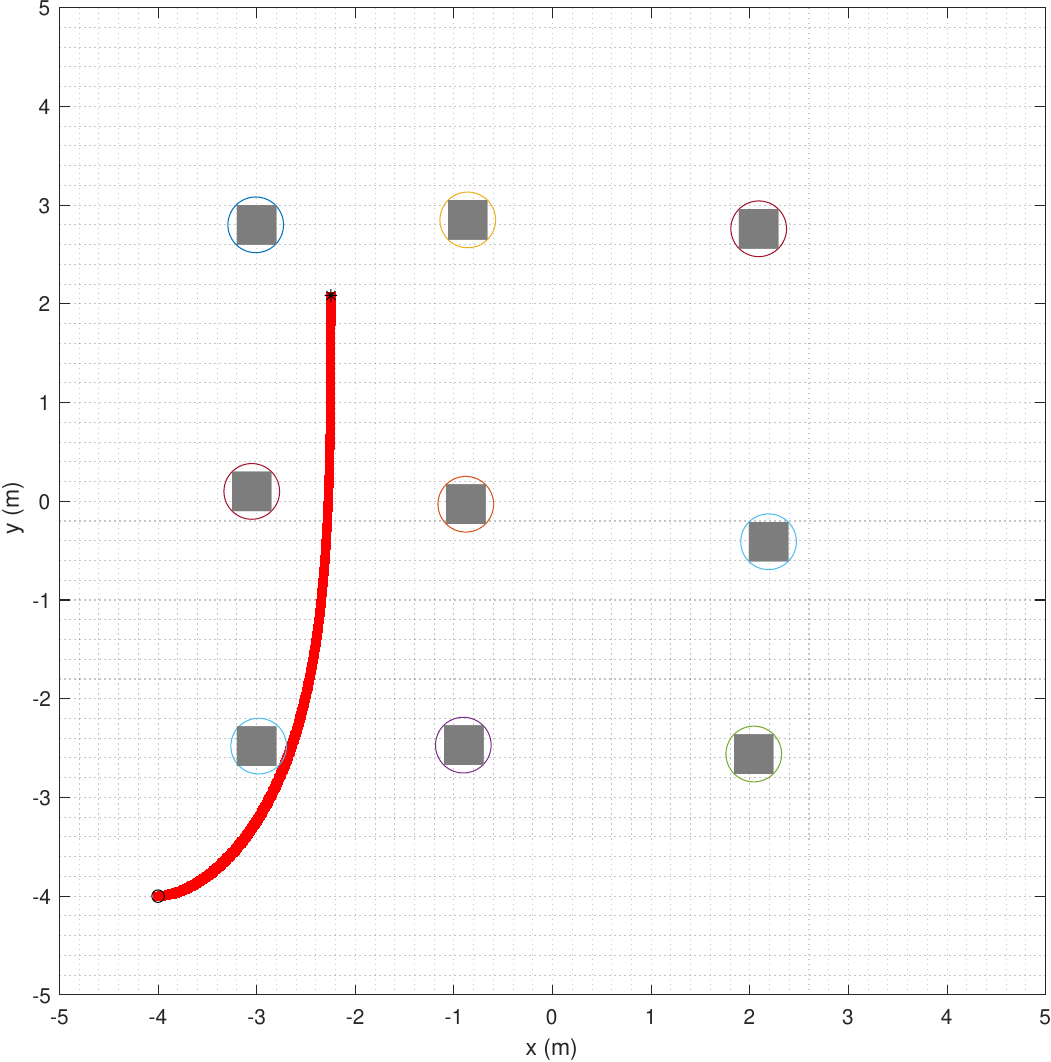}
        \caption{CLF-CBF-QP-SRP for target point 3}
        \vspace{0.2cm}
        \label{fig:7pdf}
    \end{subfigure}
    \hfill
    \raisebox{0.3cm}{
        \begin{subfigure}[b]{0.3\textwidth}
            \includegraphics[width=\textwidth]{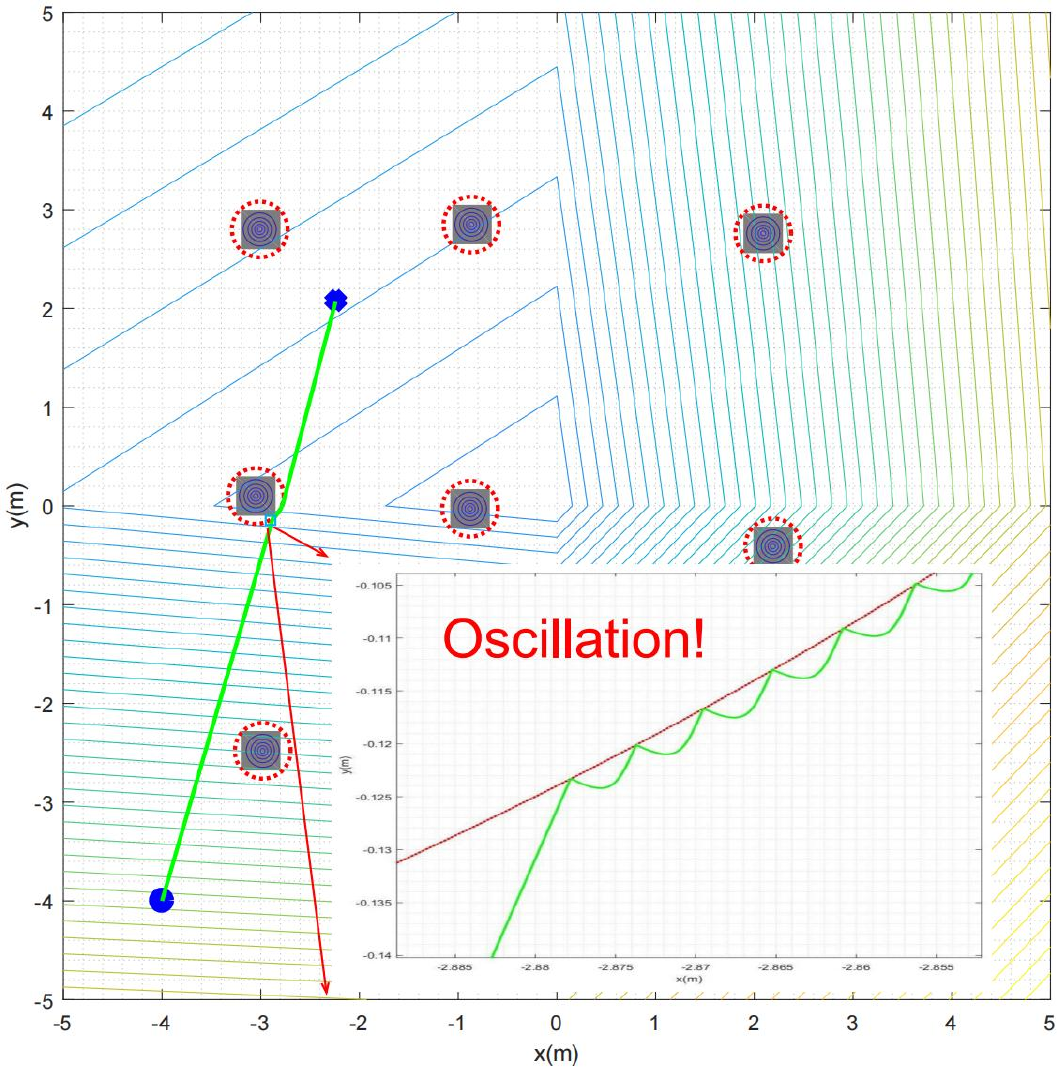}
            \caption{APF for target point 3}
            \vspace{0.3cm}
            \label{fig:8jpg}
        \end{subfigure}
    }
    \hfill
    \begin{subfigure}[b]{0.3\textwidth}
        \includegraphics[width=\textwidth]{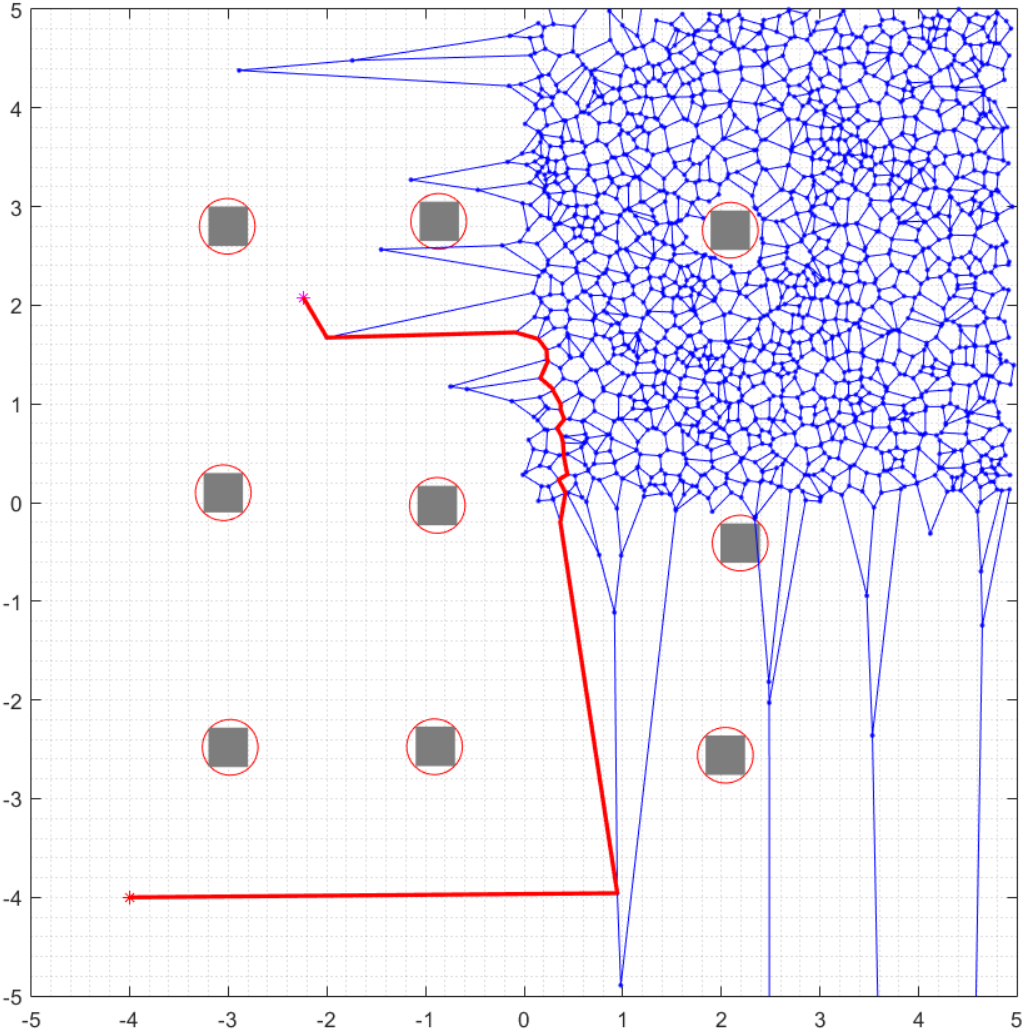}
        \caption{Voronoi diagram for target point 3}
        \vspace{0.2cm}
        \label{fig:9pdf}
    \end{subfigure}

    \caption{Comparison of CLF-CBF-QP-SRP, APF, and Voronoi diagram for three target points.}
    \label{fig:10}
\end{figure}

\subsection{Stability  and Safety of Trajectory Generation}
The stability and safety of the trajectory generation for the moving car
using CLF-CBF-QP-SRP are validated in this section through different testing target points. As shown in Fig.~\ref{figurelabel7}, three different testing scenarios are illustrated. Target point 1, target point 2 and target point 3 are located at the lower right, upper right and upper left of the centre point, respectively. These three target points are defined to test the capability of achieving stability and safety of trajectory generation with different involved obstacles. These target points are also able to test the adaptivity of the proposed CLF-CBF-QP-SRP program. In each case, different states of the moving car will be discussed.

\subsubsection{Safe and efficient trajectory generation for three target points}
Fig.~\ref{figurelabel8} shows the trajectory generation using LBF-CBF-QP-SRP program from starting point to three target points respectively. Fig.~\ref{figurelabel8}(a) illustrates the real environment for trajectory generation. Fig.~\ref{figurelabel8}(e) illustrates the detected environment by our perception. Fig.~\ref{figurelabel8}(b) illustrates the LBF-CBF-QP-SRP based global trajectory for target point 1. Fig.~\ref{figurelabel8}(c) illustrates the LBF-CBF-QP-SRP based global trajectory for target point 2 in MATLAB. Fig.~\ref{figurelabel8}(d) illustrates the LBF-CBF-QP-SRP based global trajectory for target point 3 in MATLAB. Fig.~\ref{figurelabel8}(f) illustrates the final trajectory with local safe planning for target point 1 in Rviz. Fig.~\ref{figurelabel8}(g) illustrates the final trajectory with local safe planning for target point 2 in Rviz. Fig.~\ref{figurelabel8}(h) illustrates the  final trajectory with local safe planning for target point 3 in Rviz. From these trajectories, it can be observed that the moving car can avoid collisions and drive close to the boundary of reshaped obstacles, improving the efficiency of reaching the target point. Besides, all the trajectories are smooth and stable, as there is no sharp changing. Therefore, the proposed LBF-CBF-QP-SRP program can ensure the safety, smoothness and stability of driving among the whole process.

\subsubsection{Safe and stable control of target point 3}
This section illustrates the states, control variables, and total CBF of all obstacles, from the starting point to the target point, among the time axis. As illustrated in Fig.~\ref{figurelabel9}(a), the total CBF is always larger than 0 throughout the whole time, which suggests it is safe driving. Fig.~\ref{figurelabel9}(b) shows the angular velocity of the moving car. It is obvious that the changing of the angular velocity is smooth and with a small scale, which suggests stability during the driving. Fig.~\ref{figurelabel9}(c) elaborates on the longitudinal position $x$, lateral position $y$ and course angle $\theta$\, of a moving car. The longitudinal distance has changed with smoothness gradually from -4 to the point near 0, suggesting the smoothness and stability of longitudinal driving. The lateral distance has gradually increased with smoothness from -4 to the point near 2, suggesting the smoothness and stability of lateral driving. The variation of $\theta$\, is also smooth and reaches a stable level at the end of time, revealing the stability of turning among the whole process.

\subsubsection{Comparison with Advanced Trajectory Generation Algorithm}
To evaluate the performance of the LBF-CBF-QP-SRP program and enhance the validity of the simulation results, we compare it with two benchmark trajectory generation methods: Voronoi diagram \cite{jogeshwar2022algorithms} and APF \cite{mohamed2023artificial}. As discussed in Section II.C, these methods have the advantages of safety and efficiency. Therefore, we use them as basepoints for comparison. Fig.~\ref{fig:10} shows the trajectories generated by the three methods for three different target points, starting from the same initial point $(-4,-4)$. For target point 1, the LBF-CBF-QP-SRP program produces a smooth and collision-free trajectory. However, the APF method exhibits severe oscillation when it approaches the obstacle in the middle of the third row, which indicates a lack of stability. The Voronoi diagram method collides with the leftmost obstacle in the second row. Moreover, its trajectory is neither smooth nor efficient. For target point 2, the LBF-CBF-QP-SRP program generates a smooth and efficient trajectory, as it drives close to the boundary of the map-centre obstacle. However, the APF method has two noticeable oscillations with the leftmost obstacle in the third row and the map-centre obstacle, respectively, resulting in unstable driving. The Voronoi diagram method follows a longer trajectory than the other two methods, which implies low efficiency. For target point 3, the LBF-CBF-QP-SRP program ensures a smooth, efficient, and good trajectory. The APF method encounters oscillation with the leftmost obstacle in the second row. The Voronoi diagram method makes frequent turns in some parts of the driving, which reduces the efficiency. For the computational time, we test each case for ten times and calculate the average time. The average computational time of LBF-CBF-QP-SRP, APF and Voronoi diagram are 2.0176 s, 2.2723 s and 0.0732 s, respectively. Therefore, the computational time for planning the global trajectory of the proposed method is relatively low.

\section{CONCLUSION}
In this paper, we introduced an enhanced visual SLAM-based collision-free driving framework for lightweight autonomous vehicles. We employed the advanced ORB-SLAM3 algorithm, augmented with optical flow techniques to efficiently cull outliers, thereby significantly enhancing the perception capabilities of a single RGB-D camera in complex indoor environments. Our novel path planning algorithm integrates control Lyapunov function (CLF) and control barrier function (CBF) within a quadratic programming (QP) framework, which is further refined through an obstacle shape reconstruction process (SRP). The simulation experiments conducted in the Gazebo environment demonstrated that our method effectively generates safe, stable, and efficient trajectories, outperforming existing approaches in computational efficiency and trajectory optimization. The adoption of a camera-based system not only reduces reliance on heavier, more expensive sensor setups but also offers a cost-effective solution with broad applicational potential in autonomous driving technologies. Future efforts will focus on enhancing the adaptability of this system to dynamic environments and integrating advanced machine learning techniques to improve decision-making processes in varying scenarios.


\bibliographystyle{IEEEtran}
\bibliography{IEEEabrv,zq_lib}

\end{document}